\documentclass{article}

\usepackage[final]{neurips_2019}

\usepackage[utf8]{inputenc}
\usepackage[T1]{fontenc}
\usepackage{hyperref}
\usepackage{url}
\usepackage{booktabs}
\usepackage{amsfonts}
\usepackage{nicefrac}
\usepackage{microtype}
\usepackage{graphicx}
\usepackage{xcolor}
\usepackage{lipsum}

\usepackage{amsmath}
\usepackage{subcaption}

\title{
  Transfer Learning for FHIR Questionnaire Terminology Binding
}

\author{
  Maxim Gorshkov \\
  Department of Computer Science \\
  Stanford University \\
  \texttt{maximgor@stanford.edu} \\
}

\begin{document}

\maketitle

\begin{abstract}
Electronic prior authorization workflows require FHIR Questionnaire items to carry
LOINC codes, yet most items in the HL7 Da Vinci CDS-Library lack these bindings.
We treat this as a retrieval problem: given a Questionnaire item's text, find the
correct LOINC code in a pool of 97,314 active codes.
We compare six methods (TF-IDF, frozen MiniLM, BioBERT, BioLORD, contrastively
fine-tuned MiniLM, and a TF-IDF+GPT reranker) on a 54-item evaluation set
spanning three query styles (natural question, medium, and terse).
No single method wins on every metric. BioLORD, a frozen encoder pre-trained on
biomedical ontology definitions, has the best top-rank accuracy (R@1 = 0.185,
MRR = 0.246) despite seeing no task-specific data, while a contrastive fine-tune
on raw LHC-Forms pairs takes R@5 (0.389) and R@10 (0.426).
A distribution-shift ablation shows why the fine-tune in our main table is not the
strongest one: adding GPT-generated paraphrases to the raw pairs drops R@5 from
0.389 to 0.296, so the augmented union underperforms raw-only training on every
metric except R@1.
Performance peaks at 5k training pairs.
Error analysis on BioLORD's R@1 failures shows that wrong-specificity and
ambiguous-text cases together account for 59\% of errors.
\end{abstract}

\section{Introduction}

FHIR Questionnaires are machine-readable forms used to drive electronic prior
authorization workflows. Each question in a Questionnaire needs a standardized
code (typically from LOINC) so that the downstream form prepopulation engine 
can find the matching data in a patient's EHR. Without that code, the
form is syntactically valid but the prepopulation engine has no way to know what to look up.

In practice, most Questionnaires lack these bindings. Across 12 substantive
rulesets in the HL7 Da Vinci CDS-Library \citep{cdslibrary}, a reference
implementation for prior authorization \citep{dtrig}, only 39 of 702 codable
items (5.6\%) carry explicit \texttt{item.code} bindings, and only 21 of those
are real LOINC codes. Authoring tools currently leave the binding step for a
human.

This project frames the problem as retrieval: given a Questionnaire item's text,
find the correct LOINC code in a fixed candidate pool. We train on (item text,
LOINC code) pairs from the NLM LHC-Forms FHIR service \citep{lhcforms}, a
collection of 3,413 Questionnaires with near-complete LOINC coverage, and
evaluate on CDS-Library items with confirmed LOINC codes.

The core difficulty is a style mismatch: LHC-Forms texts are in LOINC
concept-name style (e.g., ``Breath H2 pre carb'', ``Weight''), whereas CDS-Library items
are clinician-authored questions (e.g., ``Was arterial blood gas study ordered
and evaluated?''). A retriever trained on terse concept names does not
automatically transfer to natural-question queries.

Our contributions are: (1) a 54-item evaluation set assembled from three CDS-Library
extraction methods, including 28 items traced through CQL prepopulation expressions;
(2) a comparison of six retrieval approaches across three query styles; and (3) an
analysis showing that domain pre-training wins top-rank accuracy (R@1 and MRR)
while a fine-tune on raw in-domain pairs wins recall at 5 and 10, and that
paraphrase augmentation degrades performance when the training and evaluation data
share lexical structure.

\section{Related work}

\paragraph{Bi-encoder retrieval.}
\citet{karpukhin-etal-2020-dense} introduced dense passage retrieval, encoding queries and
documents into a shared embedding space and ranking by dot product.
\citet{sbert} showed that BERT-style encoders need contrastive sentence-level
training for meaningful cosine-similarity rankings.

\paragraph{Biomedical encoders.}
BioBERT \citep{biobert} adapts BERT via continued pre-training on PubMed and PMC.
BioLORD \citep{biolord} grounds representations in ontology definitions and
knowledge graph structure, which our results suggest is well-suited to LOINC.

\paragraph{Domain adaptation and LLMs.}
GPL \citep{gpl} synthesizes pseudo-queries to fine-tune dense retrievers without
manual annotation; our paraphrase augmentation follows the same idea. On the LLM
side, \citet{fhirgpt} show that GPT-based models can map clinical text to FHIR
resources, but their setting involves extraction into structured fields rather
than retrieval over a 97k-code vocabulary.

\section{Methods}

\subsection{Data}

\paragraph{Candidate pool.}
We use the full LOINC release: 97,314 active codes, each represented
by its \texttt{LONG\_COMMON\_NAME} display string. This is the realistic retrieval
setting, with no prior filtering on likely codes.

\paragraph{Training data.}
We extract (item text, LOINC code) pairs from the NLM LHC-Forms FHIR server,
producing 97,430 pairs across 3,413 Questionnaires. The same concepts recur across
many Questionnaires, so we deduplicate on the (item text, LOINC code) key, which
leaves 30,856 unique pairs spanning 29,746 distinct LOINC codes.
Because LHC-Forms Questionnaires are generated directly from LOINC panel
definitions, 99.99\% of item texts are identical to the LOINC display name. A
model trained naively on these pairs learns to discriminate between LOINC concepts
but does not learn to bridge from natural-question phrasing to LOINC vocabulary.
To address this, we use GPT-4o-mini to produce three paraphrases per unique pair
(e.g., from ``Body weight'' to ``What is the patient's current weight?''),
following the spirit of GPL \citep{gpl}. This adds 92,564 paraphrase pairs, for a
combined training set of 123,420 (30,856 raw + 92,564 paraphrase).

\paragraph{Evaluation set.}
We build a 54-item evaluation set from the CDS-Library by combining three
extraction methods: 21 items with an explicit \texttt{item.code} using the LOINC
system URL, 5 items whose \texttt{linkId} encodes a LOINC code
(e.g., \texttt{/20564-1-current}), and 28 items traced through CQL prepopulation
expressions to their referenced LOINC codes. We label each item by query style:
items ending in ``?'' are natural questions ($n=6$), items with at most 4 tokens
are terse ($n=34$), and the rest are medium ($n=14$). Most CDS-Library items are
terse, which is the style closest to LHC-Forms training data.

\subsection{Retrieval methods}

\paragraph{TF-IDF.}
Word-level unigram and bigram TF-IDF over LOINC display names, ranked by cosine
similarity. This is the lexical baseline and a natural lower bound.

\paragraph{Frozen MiniLM.}
The \texttt{all-MiniLM-L6-v2} sentence encoder \citep{sbert}, pre-trained on
general-domain text pairs using a contrastive objective.

\paragraph{Frozen BioBERT.}
\texttt{dmis-lab/biobert-v1.1} \citep{biobert}, mean-pooled over last hidden
states.

\paragraph{Frozen BioLORD.}
\texttt{FremyCompany/BioLORD-2023} \citep{biolord}, pre-trained to align
biomedical concept names with their ontology definitions and knowledge graph
neighbors.

\paragraph{Contrastive fine-tune (FT).}
We fine-tune \texttt{all-MiniLM-L6-v2} on LHC-Forms pairs using Multiple Negatives
Ranking Loss (MNRL). MNRL treats all other items in the batch as negatives and is
efficient for large candidate pools. Training uses batch size 64, learning rate
$2\!\times\!10^{-5}$, and 3 epochs. Unless otherwise noted, training data is the
union of raw and paraphrase pairs (123,420 total).

\paragraph{LLM reranker.}
TF-IDF retrieves the top 50 candidates, and GPT-4o-mini re-ranks them via a zero-shot
prompt.

\section{Experiments}

\subsection{Main results}

Table~\ref{tab:main} reports retrieval performance for all six methods.
Among these six, BioLORD has the best R@1 (0.185), R@5 (0.315), and MRR (0.246),
so a frozen domain encoder with no knowledge of our task leads a model trained on
in-domain data on three of four metrics.
The contrastive fine-tune in this table is trained on the union of raw and
paraphrase pairs; it wins R@10 outright (0.370 versus BioLORD's 0.333), placing
the gold code in the top 10 more often than any other method.
The ordering is not as clean as a single table makes it look: the raw-only
fine-tune in Table~\ref{tab:e2} reaches R@5 = 0.389 and R@10 = 0.426, beating
BioLORD on both, so BioLORD's lead holds for top-rank accuracy but not for recall
at 5 and 10. With only 54 evaluation items these margins are small (one item moves
R@5 by 1.9 points), so we read the table as a coarse ordering rather than a precise
ranking.
The LLM reranker leads on R@1 (0.130) among the fine-tuned methods and is
competitive with contrastive FT overall, but loses to BioLORD at the top.

\begin{table}[t]
  \centering
  \small
  \begin{tabular}{lcccc}
    \toprule
    Method              & R@1   & R@5   & R@10  & MRR   \\
    \midrule
    TF-IDF              & 0.093 & 0.167 & 0.222 & 0.119 \\
    MiniLM (frozen)     & 0.056 & 0.222 & 0.315 & 0.137 \\
    BioBERT (frozen)    & 0.056 & 0.056 & 0.074 & 0.058 \\
    \textbf{BioLORD (frozen)}
                        & \textbf{0.185} & \textbf{0.315} & 0.333 & \textbf{0.246} \\
    Contrastive FT      & 0.111 & 0.296 & \textbf{0.370} & 0.178 \\
    LLM (TF-IDF+GPT)    & 0.130 & 0.241 & 0.278 & 0.181 \\
    \bottomrule
  \end{tabular}
  \caption{Retrieval performance on the 54-item CDS-Library evaluation set,
  full LOINC pool (97,314 codes). Best result in bold.}
  \label{tab:main}
\end{table}

\begin{figure}[t]
  \centering
  \begin{subfigure}[t]{0.48\linewidth}
    \includegraphics[width=\linewidth]{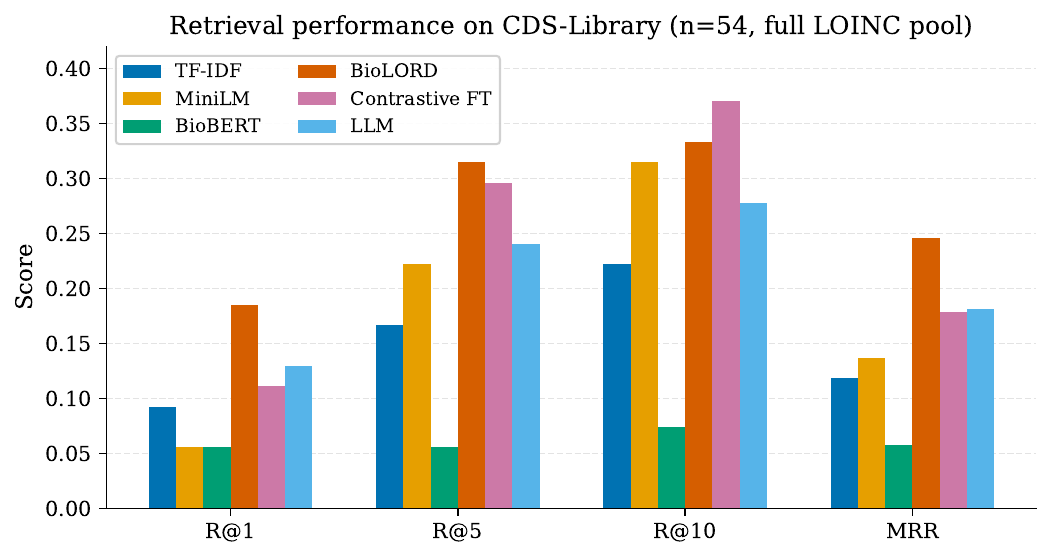}
    \caption{R@1, R@5, R@10, MRR for all methods.}
    \label{fig:f1}
  \end{subfigure}
  \hfill
  \begin{subfigure}[t]{0.48\linewidth}
    \includegraphics[width=\linewidth]{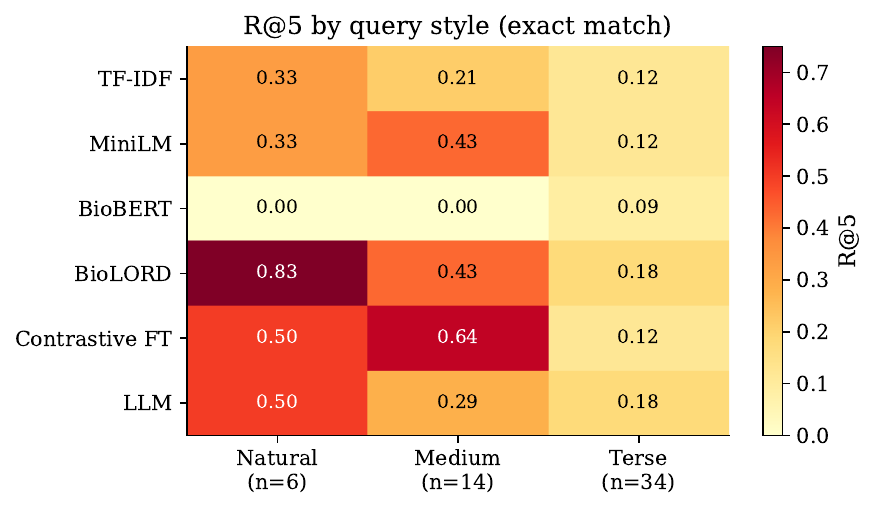}
    \caption{R@5 by query style. All methods struggle on terse items (n=34).}
    \label{fig:f3}
  \end{subfigure}
  \caption{Main results: overall performance (left) and per query style (right).}
  \label{fig:e1}
\end{figure}

A concept-family sub-metric credits a retrieval if the predicted LOINC's display
starts with the same leading token as the gold code.
BioBERT's exact R@5 is 0.056, but its family R@5 is 0.261, so it gets the right
LOINC concept family but cannot distinguish within-family variants (e.g.,
O2 saturation vs.\ O2 partial pressure in blood). The fine-tuning goal then
becomes within-family discrimination rather than learning clinical semantics
from scratch.

Figure~\ref{fig:f3} shows R@5 by query style. BioLORD is substantially stronger
on natural-question queries than any other method. Contrastive FT leads on
medium-length items. Every method struggles on terse items (63\% of the eval
set), where the distribution shift from training data is smallest but specificity
errors dominate.

\subsection{Distribution-shift ablation}

Table~\ref{tab:e2} compares three training variants of the contrastive fine-tune.
Raw-only (original LHC-Forms pairs, no paraphrases) is the strongest variant on
every metric except R@1: R@5 = 0.389 and R@10 = 0.426, the latter being the
highest R@10 of any model or variant in this paper.
The paraphrase-only model lands in the middle (R@5 = 0.315), and the union of
both is the weakest at R@5 = 0.296.

\begin{table}[t]
  \centering
  \small
  \begin{tabular}{lcccccc}
    \toprule
    Training & R@1 & R@5 & R@10 & MRR & Nat.\ R@5 & Terse R@5 \\
    \midrule
    Raw only    & 0.074 & \textbf{0.389} & \textbf{0.426} & \textbf{0.202} & \textbf{0.667} & \textbf{0.235} \\
    Para.\ only & \textbf{0.130} & 0.315 & 0.389 & 0.192 & 0.500 & 0.176 \\
    Union       & 0.111 & 0.296 & 0.370 & 0.178 & 0.500 & 0.118 \\
    \bottomrule
  \end{tabular}
  \caption{Distribution-shift ablation ($n=54$). Raw only trains on 30,856
  deduplicated pairs, Para.\ only on 92,564 paraphrases, and Union on all 123,420.
  ``Nat.'' and ``Terse'' are per-style R@5 for natural-question ($n=6$) and terse
  ($n=34$) items. ``Union'' is the method in Table~\ref{tab:main}.}
  \label{tab:e2}
\end{table}

We added paraphrases to bridge the terse-to-natural style gap, yet the union is
worse than raw alone at every metric except R@1. Two explanations seem plausible. First, the LOINC candidate pool uses terse
display names, so a model trained on terse pairs is naturally calibrated to the
query-to-pool style alignment; adding paraphrases shifts representations away
from that alignment. Second, MNRL treats all other items in a batch as negatives,
so when a paraphrase and the original of the same item share a batch, they become
negatives for each other despite sharing the same gold LOINC code.

\subsection{Training size ablation}

\begin{figure}[t]
  \centering
  \includegraphics[width=0.55\linewidth]{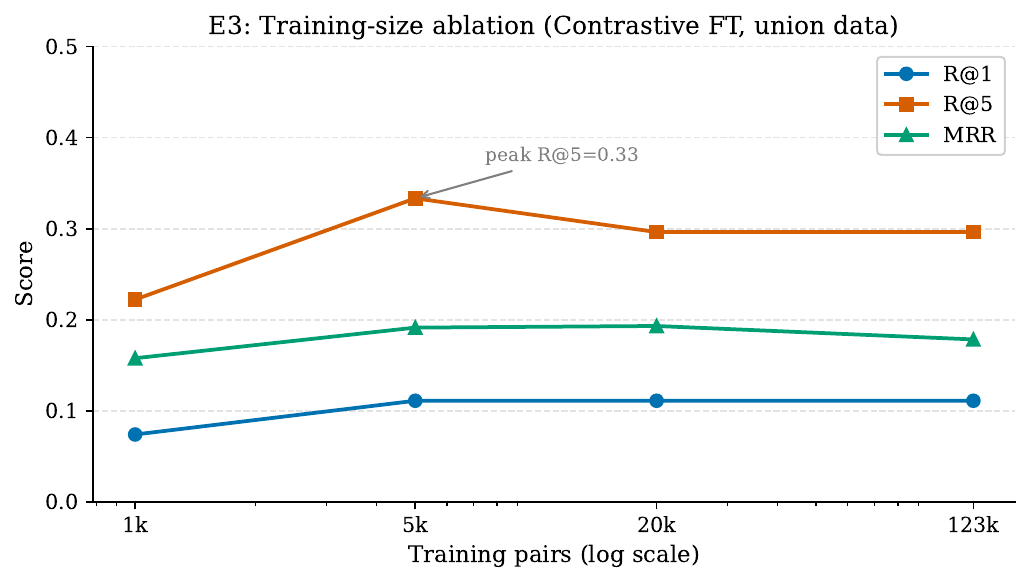}
  \caption{R@5 vs.\ training set size (union data).
  Performance peaks at 5k pairs.}
  \label{fig:f6}
\end{figure}

Figure~\ref{fig:f6} shows R@5 as a function of training set size.
Performance improves from 1k (R@5 = 0.222) to 5k (R@5 = 0.333), then flat at
20k and the full 123k (both 0.296).
The marginal pairs beyond 5k are mostly additional paraphrases of the same items,
so they do not add new signal.

\subsection{Error analysis}

\begin{figure}[t]
  \centering
  \includegraphics[width=0.75\linewidth]{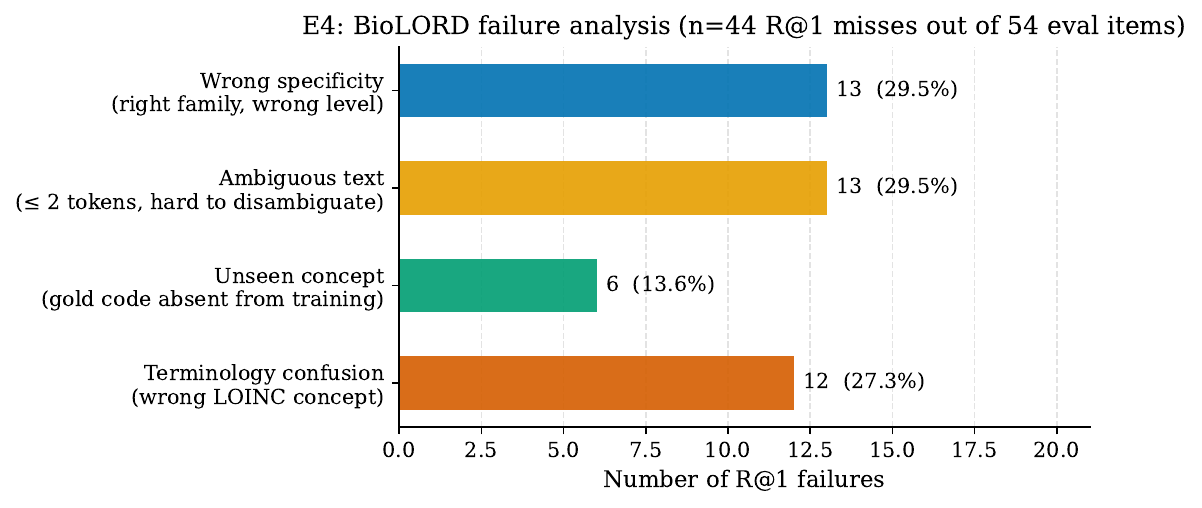}
  \caption{BioLORD R@1 failure breakdown ($n=44$). Seven rows were
  manually corrected from the initial output of the error-bucketing heuristic script.}
  \label{fig:f7}
\end{figure}

We examined the 44 BioLORD items where R@1 failed and assigned each to one of
four buckets. Figure~\ref{fig:f7} shows the breakdown.

\textit{Wrong specificity} (29.5\%): BioLORD retrieves a LOINC code from the right
concept family but the wrong member, usually because the specimen type or
measurement method differs (e.g., arterial vs.\ whole blood for pH).

\textit{Ambiguous text} (29.5\%): The item text is very short (one or two tokens)
and maps to several plausible codes with no further context to disambiguate.
``PH'' could be pH in arterial blood (2744-1) or pH in whole blood (11558-4).

\textit{Terminology confusion} (27.3\%): The prediction is from a completely
different LOINC concept. For ``PaCO2 [mmHg]'', the gold code is
\texttt{2019-8} (\emph{Carbon dioxide [Partial pressure] in Arterial blood}),
but BioLORD returned \texttt{38645-8} (\emph{Captan [Mass/volume] in Air}), which is a
pesticide measurement with no clinical overlap with the query.

\textit{Unseen concept} (13.6\%): The gold LOINC code does not appear in the
LHC-Forms training vocabulary.

Specificity and ambiguity together account for 59\% of failures. The specificity
errors in particular suggest that within-family discrimination is the remaining bottleneck.

\section{Conclusion}

BioLORD's domain pre-training on biomedical ontologies gives it the best top-rank
accuracy: it wins R@1 and MRR despite never seeing our training pairs. However,
the raw-only contrastive fine-tune takes R@5 (0.389) and R@10 (0.426), 
and even the union fine-tune in Table~\ref{tab:main} retrieves the gold
code into the top 10 more often than BioLORD (0.370 vs.\ 0.333). A possible
explanation is that BioLORD's training distribution aligns well with the
evaluation items that have the most clinical framing, while the fine-tuned model
learns to discriminate within the LHC-Forms vocabulary and overfits to the terse
style of its training pairs.

The raw-only model achieves R@5 = 0.389 versus 0.296 for the union,
despite seeing fewer, stylistically narrower pairs. The candidate pool is terse,
and a model trained on terse pairs is naturally calibrated to it. Adding
paraphrases shifts the query representations away from the pool vocabulary
without changing the pool.

The main limitation of this work is the small evaluation set ($n=54$). With 54
items, three correct retrievals is a 5.6 percentage point swing in R@5, so
methods that are close together in Table~\ref{tab:main} may not be meaningfully
different. A larger evaluation set, ideally from real prior-authorization
implementations, would make the rankings more reliable.

The most promising direction for future work is a cross-encoder reranker applied
to the top candidates from our raw-only fine-tune, which reaches R@10 = 0.426, meaning
the gold code is in its top 10 for 43\% of queries. A cross-encoder that sees both the 
query and candidate together should handle the within-family specificity errors that 
account for nearly a third of current failures. Incorporating LOINC hierarchy
structure into the retrieval objective would
also directly address the specificity problem.

\section*{Source code}

Source code for all experiments is available at
\url{https://github.com/maximg0r/fhir-questionnaire-terminology-binding}.

\bibliographystyle{acl_natbib}
\bibliography{references}

\end{document}